\documentclass[11pt,a4paper]{article}
\usepackage[hyperref]{emnlp2020}
\usepackage{times}
\usepackage{latexsym}

\usepackage{microtype}

\aclfinalcopy % Uncomment this line for the final submission
\def\aclpaperid{2963} %  Enter the acl Paper ID here

\usepackage{tabularx} % extra features for tabular environment
\usepackage{array}
\newcolumntype{L}[1]{>{\raggedright\let\newline\\\arraybackslash\hspace{0pt}}m{#1}}
\newcolumntype{C}[1]{>{\centering\let\newline\\\arraybackslash\hspace{0pt}}m{#1}}
\newcolumntype{R}[1]{>{\raggedleft\let\newline\\\arraybackslash\hspace{0pt}}m{#1}}

\usepackage[hang,flushmargin]{footmisc} % footnote indent

\usepackage{amsmath}  % improve math presentation
\usepackage{graphicx} % takes care of graphic including machinery
\usepackage{mathtools}
\usepackage{amssymb}
\usepackage{booktabs} % For formal tables
\usepackage{multirow}
\usepackage{adjustbox}
\usepackage{makecell}
\usepackage{threeparttable}
\usepackage{subfigure}
\usepackage{overpic}
\usepackage{bm} % bm
\usepackage{soul}

\usepackage{balance} % balance references
\usepackage{booktabs} % professional table
\usepackage{threeparttable} % table footnote
\usepackage{enumitem} % enumerate

\newcommand{\RNum}[1]{\uppercase\expandafter{\romannumeral #1\relax}}

\newsavebox\CBox

\title{Generalizable and Explainable Dialogue Generation via \\ Explicit Action Learning}

 \author{Xinting Huang,\textsuperscript{1}
Jianzhong Qi,\textsuperscript{1}
Yu Sun,\textsuperscript{2}
Rui Zhang\textsuperscript{1}\Thanks{ Rui Zhang is the corresponding author.}\\
\textsuperscript{1}{The University of Melbourne}, 
\textsuperscript{2}{Twitter Inc.}\\
\{xintingh@student., jianzhong.qi@, rui.zhang@\}unimelb.edu.au,
ysun@twitter.com}

\date{}

\begin{document}
\maketitle
\begin{abstract}

Response generation for task-oriented dialogues implicitly optimizes two objectives at the same time: task completion and language quality.
Conditioned response generation serves as an effective approach to separately and better optimize these two objectives. 
Such an approach relies on system action annotations which are expensive to obtain.
To alleviate the need of action annotations, latent action learning is introduced to map each utterance to a latent representation.
However, this approach is prone to over-dependence on the training data, and the generalization capability is thus restricted.  
To address this issue, we propose to learn \emph{natural language actions} that represent utterances as a span of words. 
This explicit action representation promotes generalization via the compositional structure of language.
It also enables an explainable generation process.
Our proposed unsupervised approach learns a memory component to summarize system utterances into a short span of words.
To further promote a compact action representation, we propose an auxiliary task that restores state annotations as the summarized dialogue context using the memory component.
Our proposed approach outperforms latent action baselines on MultiWOZ, a benchmark multi-domain dataset.

\end{abstract}

\section{Introduction}

Task-oriented dialogue systems complete tasks for users, such as making a hotel reservation or finding train routes, in a multi-turn conversation \cite{gao2018neural,sun2016contextual,sun2017collaborative}.
The generated system utterances should not only be naturally sound, but more importantly be \emph{informative}, i.e., to proceed the dialogue towards task completion.
To fulfill this requirement, conditioned response generation is widely adopted based on system actions \cite{wen2017latent,chen2019semantically}. 
The response generation process is decoupled into two consecutive steps, where an action is first selected and then an utterance is generated conditioned on this action.
One can optimize each step towards its goal, i.e., informative and naturally sound, without impinging the other \cite{yarats2018hierarchical}.
However, such approaches rely on action annotations (as in Table \ref{toy-example}), which require domain knowledge and extensive efforts to obtain.

\begin{table} [!tbp]
    \centering
\begin{threeparttable}
    \centering
\caption{ \small Conditioned Response Generation Example 
} \label{toy-example}
% \tiny
%\scriptsize
%\footnotesize
\small
% \normalsize  
\setlength\tabcolsep{3.0pt}
\begin{tabular}{L{1.5cm}|L{1.8cm}|L{3.8cm}}
\toprule
% \multicolumn{2}{c}{\textbf{System utterances}} \\
\multirow{4}{*}{\makecell[l]{Dialog\\Context} 
} &  \multicolumn{2}{l}{\textbf{Utterance}}\\
\cmidrule{2-3}
& \multicolumn{2}{L{5.8cm}} {User: \textit{I need a train that departs bishops stortford on wednesday, please.} }\\ 
\cmidrule{2-3}
& \multicolumn{2}{l} {\textbf{Dialogue state annotation} }\\ 
\cmidrule{2-3}
& \multicolumn{2}{L{5.9cm}}{\texttt{Train:\{departure=bishops stortford, day=wednesday\}}    }\\
\midrule
\midrule
\multirow{5}{*}{\makecell[l]{Action\\Selection} 
} & \multicolumn{2}{l} {\textbf{System action annotation} }\\ 
\cmidrule{2-3}
& \multicolumn{2}{L{5.9cm}}{ \texttt{train-inform:\{choice=five\}}; \texttt{train-request:\{leaveat\}} }\\
\cmidrule{2-3}
& \textbf{Latent action} & \makecell[l]{\textbf{Natural language Action}}  \\
\cmidrule{2-3}
&  \multirow{2}{*}{\textit{[0,0,0,1,0]}}  & \textit{\{`option', `five' }  \\ 
& & \textit{`request',`time' \}}\\
\midrule
\makecell[l]{Response\\Generation} & \multicolumn{2}{L{5.9cm}}{ \makecell[l]{\textit{alright I found five options available. when } \\ \textit{would you like to leave by?} } }\\
% \multirow{2}{*}{\makecell[l]{Response\\Generation}} & \multicolumn{2}{l}{\textbf{System response}}  \\
% \cmidrule{2-3}
% & \multicolumn{2}{L{5.9cm}}{ \makecell[l]{\textit{alright I found five. when would you like } \\ \textit{to leave by?} } } \\
\bottomrule
\end{tabular}
% \begin{tablenotes}\footnotesize
% \item[*]
% \end{tablenotes}
\end{threeparttable}
\end{table}

To deal with the absence of action annotations, latent action learning has been introduced \cite{zhao2018unsupervised,yarats2018hierarchical}. 
System utterances are represented as low-dimensional latent variables by an auto-encoding task \cite{zhao2019rethinking}, and utterances with the same representations are considered to convey similar meanings. 
Such action representations might be prone to over-dependence on the training data, which restricts the model generalization capability, especially when multiple domains are considered.
% This is because the \emph{implicit} nature of latent variables makes it unable to enforce the desired properties of the latent space, i.e., to capture the intentions of system utterances, without explicit supervision \cite{Locatello2019ChallengingCA}. 
This is because, without explicit supervision, the desired property of capturing the intentions of system utterances in the latent space cannot be enforced \cite{Locatello2019ChallengingCA}, which in turn is due to the implicit nature of latent variables.
For example, variational auto-encoder (VAE), which is often used for latent action learning, tends to produce a balanced distribution over the latent variables \cite{zhao2018unsupervised}, while the true distribution of system actions is highly imbalanced \cite{Budzianowski2018MultiWOZA}.
The resulting misaligned action representations would confuse the model of both steps and degenerate the sample efficiency in training.

% This is because without explicit supervision the desired property of capturing the intentions of system utterances in the latent space cannot be enforced (L 2019), which in turn is due to the implicit nature of latent variables.

To address the above issues, we propose to learn \emph{natural language actions} that represent system utterances as a span of words, which explicitly reveal the underlying intentions.
% benefits of natural language actions
Natural language provides unique compositional structure while retaining the representation flexibility.
These properties promote model generalization and thus make natural language a ﬂexible representation for capturing characteristics with minimal assumptions \cite{jiang2019language}.
% the main rationale to obtain such actions
% In our scenarios, we aim to use language as the interface by 
Motivated by these advantages, we learn natural language actions by identifying \emph{salient words} of system utterances. 
Salient refers to indicative for a prediction task (e.g., sentiment analysis) that takes as input the original utterance.
% for the characteristics of utterances.
The main rationale is that the principal information that the task concerns can be preserved by just the salient words.
For example, the sentiment of sentence ``\textit{The movie starts out as competent but turn bland}'' can be revealed by the word ``\textit{bland}'' when it is identified salient by considering the complete context.  
In our scenarios, we consider measuring word saliency in terms of state transitions.
This is because state transitions reflect how the intentions of a system utterance influence the dialogue progress, and action representations that capture such influences can well reveal the intentions \cite{chandak2019learning,Tennenholtz2019TheNL,huang-etal-2020-semi}.
By considering salient words for state tracking tasks as actions, we obtain action representations that enjoy the merits of natural language and indeed capture the characteristics of interest, i.e., intentions of system utterances.
% explainable

% technical contributions
Obtaining salient words by applying existing saliency identification approaches \cite{Ribeiro2018AnchorsHM} is, however, unable to produce \emph{unified} action representations.
Specifically, system utterances with the same intention might not share similar wordings, and existing attribution approaches can only identify salient words within utterances.
We tackle this challenge by proposing a memory-augmented saliency approach that identifies salient words from a broader vocabulary.
The vocabulary consists of all the words that could compose natural language actions,~\footnote{We consider content words from state annotations and task descriptions, which will be specified in Sec. \ref{memory-aug}} and each word is stored as a slot in the memory component. 
By incorporating the memory component into a dialogue state tracking model, we use each system utterance as a query to perform memory retrieval, and the retrieval results are considered as salient words.
The retrieval results might contain words that are redundant since we do not have direct supervision for the retrieval operations.
For example, the resulting salient words might be ``\textit{but turn bland}'' in the example shown earlier, which include unnecessary words and may lead to degenerated action results. 
To obtain \emph{compact} action representations, we propose an auxiliary task based on pseudo parallel corpus, i.e., dialogue context and state annotation pairs. 
We observe that dialogue states serve as good examples of how compact representation should be.
Therefore, we use the encoded dialogue context as query and ask the memory component to reconstruct its text-based dialogue states.
In this way, the obtained concise actions generalize better and can be easily interpreted.

Our contributions are summarized as follows:
\begin{itemize}[topsep=0pt,leftmargin=*,noitemsep,wide=0pt]
\item
% to include unsupervised
We propose to learn explicit action representations (in contrast to latent action representations) for task-oriented dialogues, which promotes more generalizable and explainable dialogue generation.
\item
% We propose a novel memory based approach
We propose a novel memory based approach with a pseudo parallel training scheme to obtain unified and compact action representations.
\item
We conduct experiments on a benchmark multi-domain dataset.
Results show that our approach outperforms the state-of-the-art on both in-domain and cross-domain settings.
% explicit action learning
\end{itemize}

\section{Preliminaries}

Let $\{d_i|1 \leq i \leq N \}$ be a set of dialogues, and each dialogue contains $n_d $ turns: $d_i = \{(c_t, a_t,x_t)|1 \leq t \leq n_d \}$, where $c_t$ is the context at turn $t$, and $a_t$ is the dialogue action of system utterance $x_t$.
The context $c_t =\{u_1, x_1, ...,u_t \} $ consists of the dialogue history of user utterances $u$ and system utterances $x$.  
Conditioned response generation tackles the context-to-response generation problem $p(x|c)$ via two consecutive steps: a content planning step decides a dialogue action to proceed the dialogues $p_{l}(a|c)$; and a surface realization step further transforms the decided action into naturally sound utterances $p_{r}(x|a,c)$.     
% the second step: surface/language generation
% benefits
Using the two-step process, response generation could be optimized towards better task completion while maintaining high-quality language quality \cite{huang2019mala,zhao2019rethinking}.
% optimization (supervised)
The optimization process also consists of two parts. Firstly, context-action pairs are used to train the content planning model $p_{l}(a|c)$ using the cross-entropy loss.   
\begin{equation}\label{loss-content-plan}
    % \mathcal{L}=\mathbb{E}_{(c,a)}[-\log p_{l}(a|c)]    
    \mathcal{L}_{act} = \sum_{d_i}\sum_{t=1:n_d} -\log(a_t^{\top} \cdot p_{l}(a|c_t)) 
\end{equation}
Then, the surface realization model $p_{r}(x|a,c)$ is optimized from the $(c_t, a_t,x_t)$ triples to maximize the likelihood of ground-truth responses 
\begin{equation}\label{loss-nlg}
    % \mathcal{L}= \mathbb{E}_{x}[-\log p_{r}(x|c,a)] 
    \mathcal{L}_{lan} = \sum_{d_i}\sum_{t=1:n_d} -\log p_{r}(x_t|c_t,a_t)
\end{equation}
% optimization (reinforcement learning)
Furthermore, to achieve better task completion, reinforcement learning (RL) is adopted to boost the pre-trained supervised models \cite{yarats2018hierarchical,zhao2019rethinking}.
The rewards in terms of task completion (e.g., success rate) is usually computed based on the final generated response \cite{Budzianowski2018MultiWOZA}.
To avoid divergence from fluent utterances, this fine-tuning stage focuses on the content planning model $p_{l}(a|c)$ and keeps the parameters of $p_{r}(x|a,c)$ fixed.
The reward $R_{t}$ at each turn is back-propagated via policy gradients as:
\begin{equation}\label{loss-rl}
     \nabla_{\phi}\mathcal{J}(\phi) = \sum_{t=1:n_d} R_{t}\cdot\nabla_{\phi} \log p_{l}(a|c_t)
\end{equation}
% \sum_{d_i}
where $\phi$ denotes the parameters of model $p_{l}$.

% serve as effective tool to obtain meaningful and fluent responses 
% Conditioned response generation serves as an effective tool to generate meaningful and fluent system responses.
% however, annotations is rate, thus the dialogues available without a_t is more common.
% However, such approach relies on system action annotations $a_t$ which are expensive to obtain.
In order to enable conditioned response generation when action annotations are absent, latent action learning is introduced. 
Given dialogues $\{(c_t,x_t)|1 \leq t \leq n_d\}$, latent action learning aims to map each utterance to a latent representation $z_d(x)$, e.g, one-hot \cite{wen2017latent}, or continuous \cite{zhao2017learning}.
% treat latent action as the same action annotation
Based on the obtained $(c_t, z_d(x_t),x_t)$ triples, conditioned response generation is run as mentioned above.
% how the latent action is obtained
Existing latent action learning approaches mostly build on the idea of variational inference, where a latent space is found to reconstruct system utterances and thus encodes the main characteristics of utterances \cite{zhao2018unsupervised,huang2019mala}.
% suffer from the limitation -- difficult to generalize, which this study aims to address by learning.
The action representations learned from the latent space are, however, difficult to generalize due to the implicit nature and thus cause the sample inefficiency issue.

\section{Proposed Model}
\subsection{Overview}

% notations of the natural language actions and dialogues with such action representations
We study the problem of \emph{natural language action} learning for task-oriented dialogues.
Specifically, we aim to represent each system utterance $x_t$ as a sequence of word tokens $l(x_t)=[w_1,w_2,...,w_{n}]$ without dialogue action annotations.
The conditioned response generation is then performed using the enriched dialogues $\{(c_t, l(x_t), x_t)|1\leq t\leq n_d\}$. 
Since natural language actions (i.e., sequences of tokens) encode the intention of system utterances in a compact and expressive way, both dialogue planning and language generation could achieve an improved generalization capability.

% Figure 1 [\textbf{TO DO}] illustrates the overall framework of the proposed approach.
% sec 3.2 memory
We design a memory component to identify the salient words of system utterances in terms of modeling state transitions (Sec. \ref{memory-aug}).
% sec 3.3 pseudo parallel
To further boost the memory's capability in learning compact natural language actions, we propose a novel auxiliary task to identify salient words of dialogue context in a supervised setting (Sec. \ref{pseudo-para}).
% sec 3.4 conditional response generations
%  beyond the enriched dialogues, 
Furthermore,we propose to take more advantage from the action learning phase by reusing the memory component for conditioned response generation (Sec. \ref{conditional-resp}).

\subsection{Memory Augmented Action Learning}\label{memory-aug}
% to include memory component, summarize the learning objective      
We aim to obtain salient words that are indicative for the effects of system utterances in state transition.
% for state transition to form the natural language actions of system utterances.
To model the such effects, we train a dialogue state tracking model that takes as input the system utterances.
We then regard the sequence of words that substitute the system utterance and get similar state tracking results as salient words.
To obtain sequences of words (i.e., natural language actions) of such characteristics, we use a learnable memory component that stores all potential words to form action representations, and optimize the memory in a self-supervised way.

% \subsubsection*{Dialogue State Tracking}\label{subsub-dst}
% \subsubsection*{Dialogue State Tracking}
Before presenting the proposed action learning approach, we first briefly introduce dialogue state tracking tasks.
% use the notation in TRADE
For dialogues $\{(c_t,x_t)|1 \leq t \leq n_d\}$, let $\{b_t|1 \leq t \leq n_d \}$ be the dialogue state for each turn, where $b_t \in \{0,1\}^{N_b} $ and $N_b$ is the number of all slot-value pairs.
Dialogue state tracking is usually formulated as a multi-label learning problem where the state at turn $t$ predicted by modeling the conditional distribution $p(b_t|c_t)= p(b_t|u_t, x_{t-1}, b_{t-1})$, where $b_{t-1}$ is the dialogue state in the previous turn.
% and summarizes the context before the current turn.
% introduce utt encoder, context encoder and slot-value pair 
To model this conditional distribution, a state tracking model $p_{\mathcal{B}}(u_t, x_{t-1}, b_{t-1})$ mainly employs an utterance encoder, a context encoder to work with a slot-value predictor that estimates whether a slot-value pair should be included in the dialogue states \cite{lee-etal-2019-sumbt}. 
% (\textbf{TO DO:} DST model references with such components).
Specifically, the predictor takes as input a slot-value pair $(s_i,e_i)$, and the encoded utterances $h^{utt} \in \mathbb{R}^{D} $ and context $h^{ctx} \in \mathbb{R}^{D} $ from the utterance encoder $f_{utt}(u_t, x_{t-1}) $ and context encoder $f_{ctx}(b_{t-1}) $ respectively, and $D$ is the hidden dimension.
The prediction is then performed by aggregating the results of slot-value predictor $f_{val}(h^{utt}, h^{ctx}, (s_i,e_i)) $ for the complete $N_b$ slot-value pairs.
% optimization
We optimize the state tracking model using the cross-entropy loss:
\begin{equation}\label{loss-dst}
    \mathcal{L} = \sum_{d_i}\sum_{t=1:n_d}-\log(b_t^{\top} \cdot p_{\mathcal{B}}(u_t, x_{t-1}, c_{t-1}))
\end{equation}
where the parameters of $p_{\mathcal{B}}$, which include $ f_{utt}$, $ f_{ctx}$, and $ f_{val}$, are jointly trained.

Based on the learned state tracking model, a straightforward idea of obtaining salient words is to apply importance attribution approaches.  
Specifically, these approaches measure the importance of each word by observing the prediction difference caused by replacing it \cite{Ribeiro2018AnchorsHM,Jin2020TowardsHI}. 
As discussed before, this would result in different action representations for utterances with the same action.
To address this issue, we consider learning action representations from a broader vocabulary, which releases the constraint of selecting salient words only within utterances.

\subsubsection*{Key-Value Memory Component}
To this aim, we propose to use a memory component as the additional vocabulary.
Note that the selection of words to build the vocabulary is task dependent, and we select the words appearing in state annotations and content words~\footnote{We consider nouns, verbs, and adjectives as content words.} extracted from task descriptions provided in the dataset \cite{Budzianowski2018MultiWOZA}.
This simple strategy is intuitive and turns out to be empirically competitive.

Given the built vocabulary, we adopt a key-value memory bank, where each memory slot stores a word included in the vocabulary.
% the learnable parameter (key and value matrix)
Each memory slot is associated with a key vector and a value vector, given by learnable matrix $K\in \mathbb{R}^{D \times N_{v}} $ and $V \in \mathbb{R}^{D \times N_{v}}$ respectively, where $N_{v}$ is the number of words stored in the memory. 
The memory is utilized to obtain action representations by \emph{context-aware memory retrieval}. 
% to obtain $l(x_t)$, 
Specifically, we regard the encoded utterance $h_{utt}$ from the trained dialogue state tracking model as the query vector $q \in \mathbb{R}^{D} $.
% which is computed by $f_{utt}(u_t, x_{t-1}) $ from the trained dialogue state tracking model. 
The retrieval is then conducted by computing the attention weights as
\begin{equation}\label{first-hop}
    z = \text{softmax}(q^{\top} \cdot K)
\end{equation}
where $z \in \mathbb{R}^{N_v}$ is a probability vector over the slots.
% meaning of weights/probability
Memory slots with higher probability indicate that the corresponding words are expected to be more salient to represent the system utterance. 
We could assume a natural language action $l(x_{t-1})$ containing $k$ words is sampled $k$ times from a categorical distribution given by $z$ without replacement, where the value of $k$ is set as a hyper-parameter.
% In this way, we could obtain action representations based on this distribution and a determined value of $k$, e.g., set as  a hyper-parameter.

\subsubsection*{Multi-Hop Mechanism}
Building on the above sampling strategy, we further recognize that it is not plausible to assume natural language actions are of the same length by setting $k$ as a hyper-parameter. 
% This is because the expressed intentions vary across system utterances in terms of both content and amount. 
% in terms of volume
This is because the conveyed information of system utterances can vary from each other.
It is common to see certain utterances expressing more intentions than others, especially those directly determine task completion after information is accumulated.
Thus, inspired by end-to-end memory network \cite{Sukhbaatar2015EndToEndMN}, we design a multi-hop mechanism to adaptively decide the length of natural language actions.
% as Eqn \ref{first-hop}
Specifically, after obtaining the probability vector $z$, we update the query based on the original query $q$ and a weighted sum of memory values: 
\begin{equation}
    q^2 = q^{1} + z^{\top}\cdot V
\end{equation}
where $V \in \mathbb{R}^{D \times N_{v}}$ is the memory value matrix.
Note that we denote the initial query vector $q$ (i.e., $h^{utt}$)  as $q^1$ for simplicity.
% property of query q^2
Using query $q^2$, we could get a retrieval result $z^2$ as the same way in Eqn. \ref{first-hop}.

By conducting such $k$-hop memory operation (i.e., $k$ times retrieval using corresponding updated queries), we obtain $k$ different categorical distributions.
We now assume that each word is sampled from one distribution accordingly, and the length of natural language actions is indeed the number of hops carried out. 
Thus, by adaptively deciding the number of hops, we could learn variable length natural language actions.
To this aim, we design an action gate component that predicts whether to carry out a next retrieval based on the current updated query.
We perform such prediction based on the updated query, since it aggregates the information of former query and memory slots after every retrieval operation.
% Since the updated query aggregates the former query and retrieval results, and thus can provide sufficient information of whether a further hop operation is needed. 
% it is reasonable to expect the updated query to provide sufficient information of whether a further hop operation is needed.  
More specifically, we formulate the action gate as a binary random variable $t$, and its distribution is modeled as:
\begin{equation}
    p_{gate}(t|z^1,z^2,..,z^{k-1} ) =\sigma ( G^{\top} \cdot q^k )
\end{equation}
where $ \sigma(\cdot) $ is the sigmoid function, and $G \in \mathbb{R}^{D} $ is a learnable vector.
% The way to get action representation (choose whether to continue after each step)
% Based on the multi-hop mechanism with action gate, 
In this way, we can obtain natural language actions of appropriate length, which are sampled from the distributions obtained before the action gate indicate a stop of retrieving. 
% the length of action is decided by taking into account both utterance information and memory retrieval results. 

\subsubsection*{Training}
% To include continuous relaxation, loss, and denoising DST model training.
The memory component and action gate are end-to-end trained in a self-supervised way, where the feedback is whether an utterance and its action representation lead to similar state transitions, 
We can measure such similarity using a dialogue state tracking (DST) model.
% denoising training of DST
However, a direct application of the DST model trained by Eqn. \ref{loss-dst} might be prone to attribute changes between original utterances and compact natural language actions, which results in insufficient feedback.
% This could result in insufficient feedback where the DST model predicts equally poor no matter whether the input action representation is appropriate. 
To address this issue, we adopt a denoising training strategy inspired by unsupervised machine translation \cite{Lample2018UnsupervisedMT,Lample2019MultipleAttributeTR}, and obtain a DST model that is more robust to the attribute transformation. 
Specifically, we apply a noise function $g(x)$ to the utterances, and modify the DST model training loss as:
% \sum_{t=1:n_d}
\begin{equation}\label{loss-noise-dst}
    \mathcal{L}_{dst} = \sum_{d_i}-\log(b_t^{\top} \cdot p_{\mathcal{B}}(g(u_t), g(x_{t-1}), c_{t-1}))
\end{equation}
where the noise function corrupts the input utterance by performing word drops and word order shufﬂing as specified in \citet{Lample2018UnsupervisedMT}.

With a slight abuse of notations, we use $p_{\mathcal{B}}(x_{t-1})$ to denote $p_{\mathcal{B}}(u_t, x_{t-1}, c_{t-1})$.
% i.e., the prediction output of the dialogue state tracking model.
We formulate the training loss for self-supervised task as:
% \sum_{t=1:n_d} 
\begin{equation}
\begin{aligned}
    \mathcal{L}_{mem}  = &  \sum_{d_i}\Big( -\log ( b_t^{\top} \cdot p_{\mathcal{B}}(\Tilde{l}( x_{t-1})) )
    \\ &-\text{KL}(p_{\mathcal{B}}(x_{t-1}) || p_{\mathcal{B}}(\Tilde{l}( x_{t-1})) \Big)      
\end{aligned}
\end{equation}
where KL is Kullback-Leibler divergence, and $\Tilde{l}( x_{t-1}))$ is the natural language action obtained via the memory component.
% could consider re-phrase the connection between ground-truth and prediction
This loss enforces the learned action representations to restore both the ground truth and predicted state transitions.
% relaxation
Note that the natural language actions are sampled from categorical distributions, which is non-differentiable. 
To get gradients for the memory component during back-propagation, we apply a continuous approximation, i.e., using gumbel-softmax trick instead to conduct sampling \cite{jang2016categorical}, to enable end-to-end differentiability.

\subsection{Learning with Pseudo Parallel Corpus}\label{pseudo-para}
Recall that we aim to learn natural language actions that are not only expressive but also \emph{compact}, i.e., only including words that encode system intentions.
Although the memory based approach could identify salient words from a broader vocabulary, the identified words might degenerate to the words making up most of the original utterances, which introduces redundant words into action representations.
To avoid such suboptimal scenarios, we propose a supervised auxiliary task to further regularize the memory component.   
We use the encoded context $h^{ctx}$ given by $f_{ctx}(b_t)   $ from the dialogue state tracking model as query vectors, and attempt to recover the dialogue state from the memory component.
Here, we consider word-based dialogue state representations instead of multi-hot representations, $b \in \{0,1 \}^{N_b}$.
For example, the dialogue state ``\textit{food= european, price-range=moderate}'' is transformed to a text span \textit{[`food', `european', `price-range',`moderate']}.
% We then consider this text span as the expected retrieval outputs of query $h^{ctx}$.
We then form a pseudo parallel corpus by pairing the word-based dialogue states and the corresponding encoded states as
\begin{equation*}
    \mathcal{P} = \{ \big< f_{ctx}(b_t)  ,b_t^{\text{text}}. \big> |1\leq t\leq n_d \}         
\end{equation*}
where $b_t^{\text{text}}$ is the text span for $b_t$.
We train the memory component using the pseudo parallel corpus as:
\begin{equation}
\begin{split}
    \mathcal{L}_{par} = \sum _{(h,b) \in \mathcal{P}} \big( - \log p(b_t^{\text{text}}| z^{1,2,..k(b)}) 
    \\ + \sum_{2\leq i \leq k(b) }  \text{cross-entropy}(g_i, p_{gate}(q^i))     \big)
\end{split}
\end{equation}
where $k(b) $ is the length of the text span $b^{text}_t $, and $g_i \in \{0,1\} $ indicates whether the multi-hop operation should end at step $i$, and only take value one when $i$ equals $k(b)$.  
For each pair in $\mathcal{P} $, the loss consists of two terms: the first one further guides the memory component to identify salient words; meanwhile the second term enforces the memory component to \emph{only} pick salient words and promotes action representations to remain compact.
 
The overall training objective function of the natural language action learning is:
\begin{equation}
    \mathcal{L} = \mathcal{L}_{mem} + \alpha \mathcal{L}_{par} + \beta \mathcal{L}_{dst}
\end{equation}
where $\alpha $ and $\beta $ are hyper-parameters.
% Note that the dialogue state tracking model is first trained using $\mathcal{L}_{dst}$ (Eqn. \ref{loss-noise-dst}) before we conduct action learning.
The reason we include the term $\mathcal{L}_{dst}$ during action learning is to ensure the DST model provides sufficient supervision.
Some components in the DST model (i.e., $f_{utt}$ and $f_{ctx} $ ) are updated via $ \mathcal{L}_{mem} $ and $\mathcal{L}_{par}$, and by considering $\mathcal{L}_{dst}$, we could avoid a divergence from the state tracking task.

\subsection{Conditional Response Generation}\label{conditional-resp}
After obtaining natural language actions, we enrich the dialogues as $\{(c_t, l(x_t), x_t)|1\leq t \leq n_d \} $, where $l(x_t)$ is the natural language action of utterance $x_t$.
We could then run conditioned response generation to train content planning and language generation models as Eqn. \ref{loss-content-plan}-\ref{loss-rl}.  
% benefits from (in generalizations ) : better clustering, and action connection (cite action embedding two ICMLs)
The learning efficiency can be improved by the more compact and noise-free action space.
% is expected to improve since the natural language actions capture the task-relevant intentions more properly and result in a more compact and noise-free action space.   
Moreover, the natural language actions present abundant information of correlations among actions, which allows for better generalization over actions \cite{chandak2019learning, Hu2019HierarchicalDM}. 

To further enhance the generalization capability and boost the learning efficiency, we consider re-use the memory component for conditioned response generation.
Specifically, we focus on the content planning model $p_{l}(a|c)$, which aims to decide one natural language action from the action set for response generation~\footnote{We can also tackle content planning by a generative model, and details are introduced in Sec \ref{exp-discuss}}. 
We could implement the content planning model as a network that encodes the dialogue context $c$ into a hidden state of the same dimension as query vectors in the memory component.
By using the encoded results as query for memory retrieval, we obtain a distribution given by the retrieval results.
We then select the action of highest probability determined by the obtained distribution as model output.
This fine-tuning approach could not only reduce the model complexity for content planning, but also better harvest the knowledge gained in action learning phase.  
% reduce complexity

% \subsection{Discussions}

\begin{table*} [tbp]
    \centering
\begin{threeparttable}
\caption{ Multi-Domain Joint Training Results
} \label{table-joint}
% \tiny
% \scriptsize
% \footnotesize
\small
% \normalsize  
\setlength\tabcolsep{3.5pt}
\begin{tabular}{l|l|ccc|ccc|ccc}
\toprule
% Method & nDCG & $\alpha\mbox{-}$nDCG & p$\mbox{-}$nDCG & nDCG & $\alpha\mbox{-}$nDCG & p$\mbox{-}$nDCG \\
\multicolumn{2}{l|}{} & \multicolumn{3}{c|}{20\% Training Data} & \multicolumn{3}{c|}{50\% Training Data} &  \multicolumn{3}{c}{Full Training Data} \\
\cmidrule{3-11}
\multicolumn{2}{c|}{ \textsc{Model}} & Inform & Success  &BLEU & Inform & Success &BLEU & Inform & Success  &BLEU  \\
\midrule
\multirow{2}{*}{\makecell[l]{w/o Action}  } & Seq-to-Seq & 52.4 & 44.2 &11.9 & 61.6 & 50.2 & 16.4 & 71.2 & 59.9 & 18.8 \\
& TSCP &54.8	&47.3&	12.7&	66.0&	52.7&	15.6	&76.2&	64.5&	17.2\\
\midrule
\multirow{2}{*}{\makecell[l]{Continuous \\Latent Action}  } & LaRL & 51.1&	44.0&	12.7&	63.4&	50.9&	14.3&	70.8&	60.5&	14.5\\
& MALA & 55.1&	50.5&	14.1&	72.8&	63.4&	17.4&	84.1&	73.7&	18.6\\
\midrule
\multirow{2}{*}{\makecell[l]{Discrete \\Latent Action } }& LaRL  &60.5&	51.9&	10.8&	69.2&	60.1&	13.3&	81.5&	69.2&	14.8\\
& MALA &63.5&	56.2&	11.1&	74.1&	65.0&	17.1&	85.0&	76.2&	20.1\\
\midrule
\multirow{3}{*}{Proposed} & Post-hoc  &62.8&	52.4&	13.7&	68.0&	57.9&	17.2&	75.4&	62.4&	19.6\\
& Memory-based & 64.7&	55.4&	13.6&	76.1&	70.6&	19.1&	84.9&	75.2&	20.8\\
& MASP & \textbf{70.2}&\textbf{61.8}&\textbf{14.9}&\textbf{78.7}&\textbf{71.5}&\textbf{19.4}&\textbf{88.3}&\textbf{77.1}&\textbf{21.7}\\
\bottomrule
\end{tabular}
\begin{tablenotes}\footnotesize
\item[*] Note that Post-hoc and Memory-based denotes the two variants Post-hoc Saliency and Memory-based Saliency.
% whether the baseline considers conditioned generation
\end{tablenotes}
\end{threeparttable}
\end{table*}

\section{Experiments}

To show the effectiveness of the proposed approach,  \textit{\underline{m}emory-\underline{a}ugmented \underline{s}aliency with \underline{p}arallel corpus} (MASP), we experiment on two dialogue generation settings (Sec. \ref{exp-setting}).
We compare against state-of-the-art approaches in both settings (Sec. \ref{exp-res}).
We analyze the effectiveness of MASP components under different supervision ratios, and discuss how explainable generation is achieved (Sec. \ref{exp-discuss}).

\subsection{Settings}\label{exp-setting}

% dataset
We use MultiWOZ \cite{Budzianowski2018MultiWOZA}, a multi-domain human-human conversational dataset in our experiments. 
It contains in total 8438 dialogues spanning over seven domains, and each dialogue has 13.7 turns on average.
We use the separation of training, validation and testing data as original MultiWOZ dataset.
% metrics
We use the evaluation metrics as \citet{Budzianowski2018MultiWOZA} to measure dialogue task completion, which are how often the system provides a correct entity (\textbf{Inform}) and answers all the requested information (\textbf{Success}).
We use \textbf{BLEU} \cite{papineni2002bleu} to measure the language quality of generated responses.

% implementation and hyper-parameter search (\alpha, \beta, hidden dimension d, DST model selection)
We use a three-layer transformer \cite{vaswani2017attention} with a hidden size of 128 and 4 heads as our base model for content planning and response generation, i.e., $p_{l}(a|c)$ and $p_r(a,c) $ , respectively.
We use grid search to find the best hyperparameters for the models based on validation performance, which we use a combination of Inform, Success and BLEU scores to measure. 
We choose the embedding dimensionality $d$ among \{50, 75, 100, 150, 200\}, the hyperparameters $\alpha $ and $\beta $ in [0.01, 1.0].

% para to introduce Multi-domain Joint and Cross-domain setting
We consider two settings to thoroughly evaluate the conditioned response generation: \textit{multi-domain joint training} and \textit{cross-domain response generation}.
In the first setting, we train MASP and other baselines using different sizes of the training dialogues (20\%/50\%/full), and for the tasks using 20\% or 50\% of data, the distribution of dialogues across domains are kept the same as the full training set. 
In the cross-domain setting, we adopt a leave-one-out approach to evaluate the generalization ability via a more challenging few-shot learning task. 
Specifically, we use one domain as low-resource target domain (with only 1\% of dialogues are available for training) while the others as source domains.

% the baselines (w/ and w/o action learning) and the model variants (Post-hoc saliency and Memory augmented saliency)
We compare with the following baselines that do not consider conditioned generation:
(1) \textbf{Seq-to-Seq} \cite{Budzianowski2018MultiWOZA} implemented based on transformer \cite{vaswani2017attention};
(2) \textbf{TSCP} \cite{lei2018sequicity};
and two baselines that adopt latent action learning for conditioned generation:
(3) \textbf{LaRL} \cite{zhao2019rethinking};
(4) \textbf{MALA} \cite{huang2019mala}.
% discuss discrete and continuous latent actions
Note that for these two approaches, we experiment with both discrete and continuous latent action representations.
We also compare the full model \textbf{MASP} with its two variants:
(1) \textbf{Post-hoc Saliency} obtains action representations via the importance attribution technique as \citet{Jin2020TowardsHI};
(2) \textbf{Memory-based Saliency} employs the same memory component as MASP but trained without the pseudo parallel corpus.

% \subsubsection*{Multi-domain Joint Training}
% \subsubsection*{Cross-domain Response Generation}

\subsection{Overall Results}\label{exp-res}
Table \ref{table-joint} shows that MASP outperforms baselines in the multi-domain joint training setting.
MASP achieves better dialogue task completion (measured by Inform and Success) and language quality (measured by BLEU), especially in low resource scenarios.
For example, MASP (70.2) outperforms MALA (63.5) by 10.5\% under Inform when having 20\% training data.
Meanwhile, we also find that the memory component and pseudo parallel enhanced training are essential for getting effective action representations.
For example, Post-hoc Saliency (57.9) is outperformed by a large margin compared to MALA (65.0) under Success when having 50\% training data, while MASP (71.5) achieves a performance 10\% gain over MALA. 
This validates that the unified and compact characteristics are required for natural language actions to boost conditioned generation.
We further find that the contribution of the memory component and pseudo parallel corpus may vary in different ratios of training data.
For example, the memory component brings 11.9\% and 3.0\% improvements compared to Post-hoc Saliency under Inform when the ratio is 50\% and 20\% respectively, while the pseudo parallel corpus brings 3.4\% and 8.5\% improvements compared to Memory-based Saliency.
This is largely because the memory component could easily degenerate to utterance restoration when available training data is less, and thus the regularization provided by pseudo parallel corpus is more desired.

For cross-domain setting, Table \ref{table-cross-domain} includes three representative domains (\textit{hotel}, \textit{attraction}, and \textit{train}), and the observations on other domains are consistent.~\footnote{We omit the results of Seq-to-seq and TSCP in the table which are all worse than the latent action approaches}
The results show that MASP significantly outperforms the baselines in each configuration.
For example, MASP (39.2) outperforms MALA (33.9) by 15.6\% under Inform in \textit{hotel} domain.
By comparing results of Memory-based Saliency and MALA in \textit{attraction} and \textit{train}, we find that without pseudo parallel corpus, natural language actions could still be competitive occasionally.
We will conduct a detailed analysis in the next section.  
We also find that continuous latent action approaches achieve comparable results as their discrete counterparts, while the results are opposite in the joint training setting.  
For example, MALA with continuous action (41.9) is slightly outperformed by its discrete counterparts (42.2) under Success using \textit{attraction} as target.  
This is largely because the challenging cross-domain task could result in many mis-assigned action labels, and continuous action representations can still preserve certain knowledge of similarities among actions.

\begin{table*} [tbp]
    \centering
\begin{threeparttable}
\caption{ Cross-Domain Generation Results
} \label{table-cross-domain}
% \tiny
% \scriptsize
% \footnotesize
\small
% \normalsize  
\setlength\tabcolsep{3.5pt}
\begin{tabular}{l|l|ccc|ccc|ccc}
\toprule
% Method & nDCG & $\alpha\mbox{-}$nDCG & p$\mbox{-}$nDCG & nDCG & $\alpha\mbox{-}$nDCG & p$\mbox{-}$nDCG \\
\multicolumn{2}{l|}{} & \multicolumn{3}{c|}{Hotel} & \multicolumn{3}{c|}{Attraction} &  \multicolumn{3}{c}{Train} \\
\cmidrule{3-11}
\multicolumn{2}{c|}{ \textsc{Model}} & Inform & Success  &BLEU & Inform & Success &BLEU & Inform & Success  &BLEU  \\
\midrule
\multirow{2}{*}{\makecell[l]{Continuous \\Latent Action}  } & LaRL & 26.7&	22.3&	11.4&	35.3&	30.4&	13.1&	40.3&	36.4&	13.1\\
& MALA & 31.4&	30.0&	15.8&	44.6&	41.9&	16.7&	49.2&	47.0&	17.7\\
\midrule
\multirow{2}{*}{\makecell[l]{Discrete \\Latent Action } }& LaRL  &24.1&	22.7&	9.1&	35.8&	30.0&	11.8&	43.2&	40.9&	12.8\\
& MALA &33.9&	32.3&	16.7&	45.9&	42.4&	18.1&	55.6&	53.9&	19.4\\
\midrule
\multirow{3}{*}{Proposed} & Post-hoc  & 31.0&	28.8&	14.5&	42.8	&36.3&	15.6&	49.0&	45.5	&16.3\\
& Memory-based & 34.7&	32.0&	14.6&	43.3&	40.0&	17.2&	57.6&	54.2&	18.3\\
& MASP & \textbf{39.2}&\textbf{35.1}&\textbf{17.2}&\textbf{52.5}&\textbf{47.1}&\textbf{18.6}&\textbf{59.2}&\textbf{55.9}&\textbf{19.4}\\
\bottomrule
\end{tabular}
\end{threeparttable}
\end{table*}

\begin{table}[tbp]
    \centering
\begin{threeparttable}
\caption{ Effects of Content Planning Model Design} 
\label{tab-action-model}
% \tiny
% \scriptsize
% \footnotesize
\small
\setlength\tabcolsep{4.0pt}
\begin{tabular}{l|l|ccc}
\toprule
% Method & nDCG & $\alpha\mbox{-}$nDCG & p$\mbox{-}$nDCG & nDCG & $\alpha\mbox{-}$nDCG & p$\mbox{-}$nDCG \\
\multicolumn{1}{c|}{ Action} & \multicolumn{1}{c|}{ \textsc{Planning Model}} & Inform & Success  &BLEU  \\
\midrule
\multirow{3}{*}{\makecell[l]{Post-hoc\\Saliency}} & Act-\textsc{Dec} & 54.4&	47.9&	10.6 \\
\cmidrule{2-5}
& Act-\textsc{Cls} & 60.1&	50.1&	11.0\\
& Act-\textsc{Cls} (+emb) & 62.8&	52.4&	13.7\\
\midrule
\midrule
\multirow{5}{*}{\makecell[l]{MASP }} & Act-\textsc{Dec} & 64.1&	56.9&	12.6\\
& Act-\textsc{Dec} (+mem)  &  68.0&	58.6&	14.2\\
\cmidrule{2-5}
& Act-\textsc{Cls} &  64.3&	57.7&	11.4\\
& Act-\textsc{Cls} (+emb)& 68.6&	59.3&	14.0\\
& Act-\textsc{Cls} (+mem) & \textbf{70.2}& \textbf{61.8}& \textbf{14.9}\\
\bottomrule
\end{tabular}
\begin{tablenotes}\footnotesize
\item[*] Results are in multi-domain joint training of 20\% data. 
% Note that \textit{emb} and \textit{mem} stands for action embedding and memory enhancement, respectively.
\end{tablenotes}
\end{threeparttable}
\end{table}

\subsection{Discussions}\label{exp-discuss}

\begin{figure}[!t]
\centering
\subfigure[\small{Attraction as target domain}]{
\begin{overpic}[height=3.20cm]{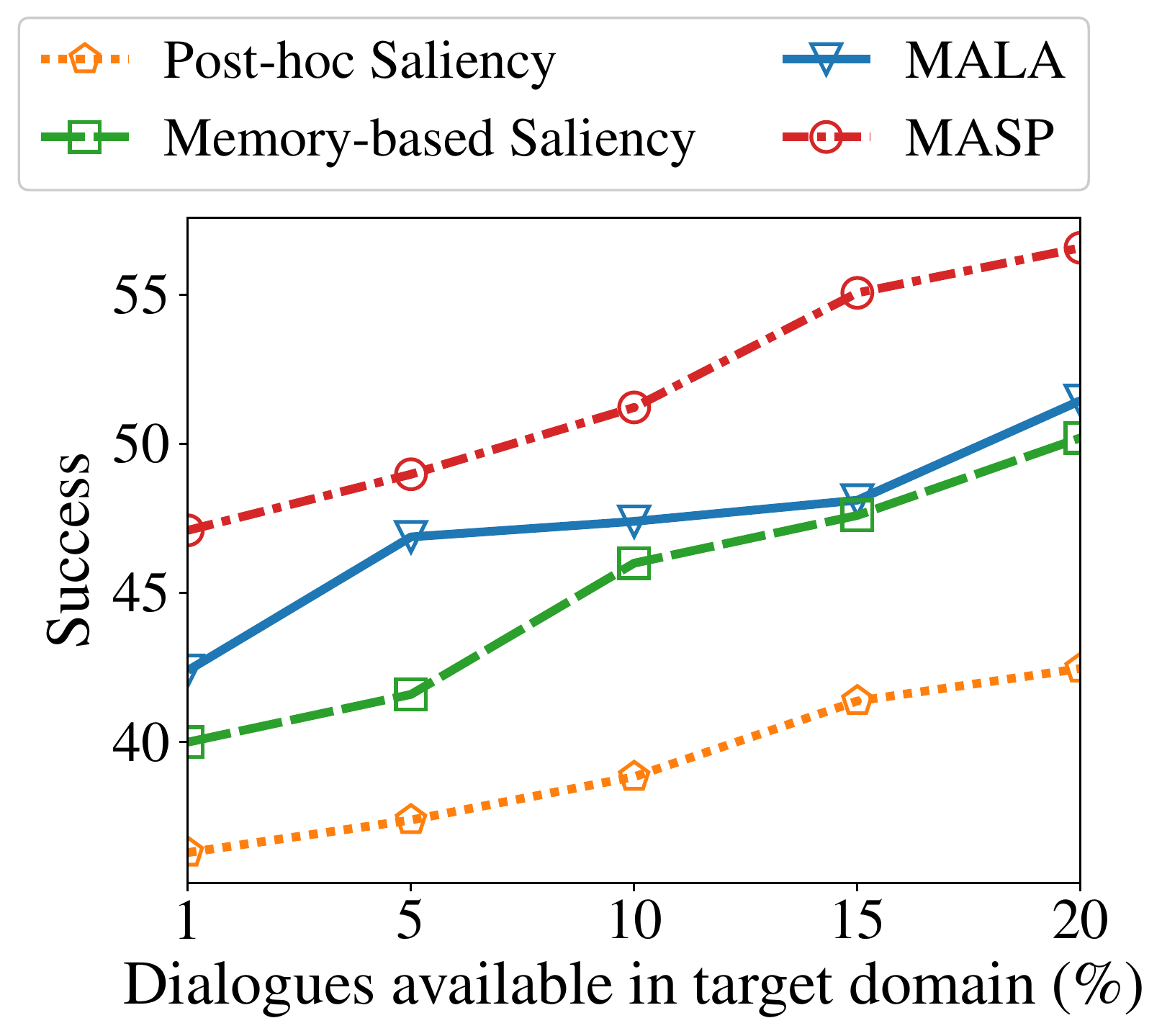}
\label{abla-target-attr}
\end{overpic}
} 
\subfigure[\small{Train as target domain}]{
\begin{overpic}[height=3.20cm]{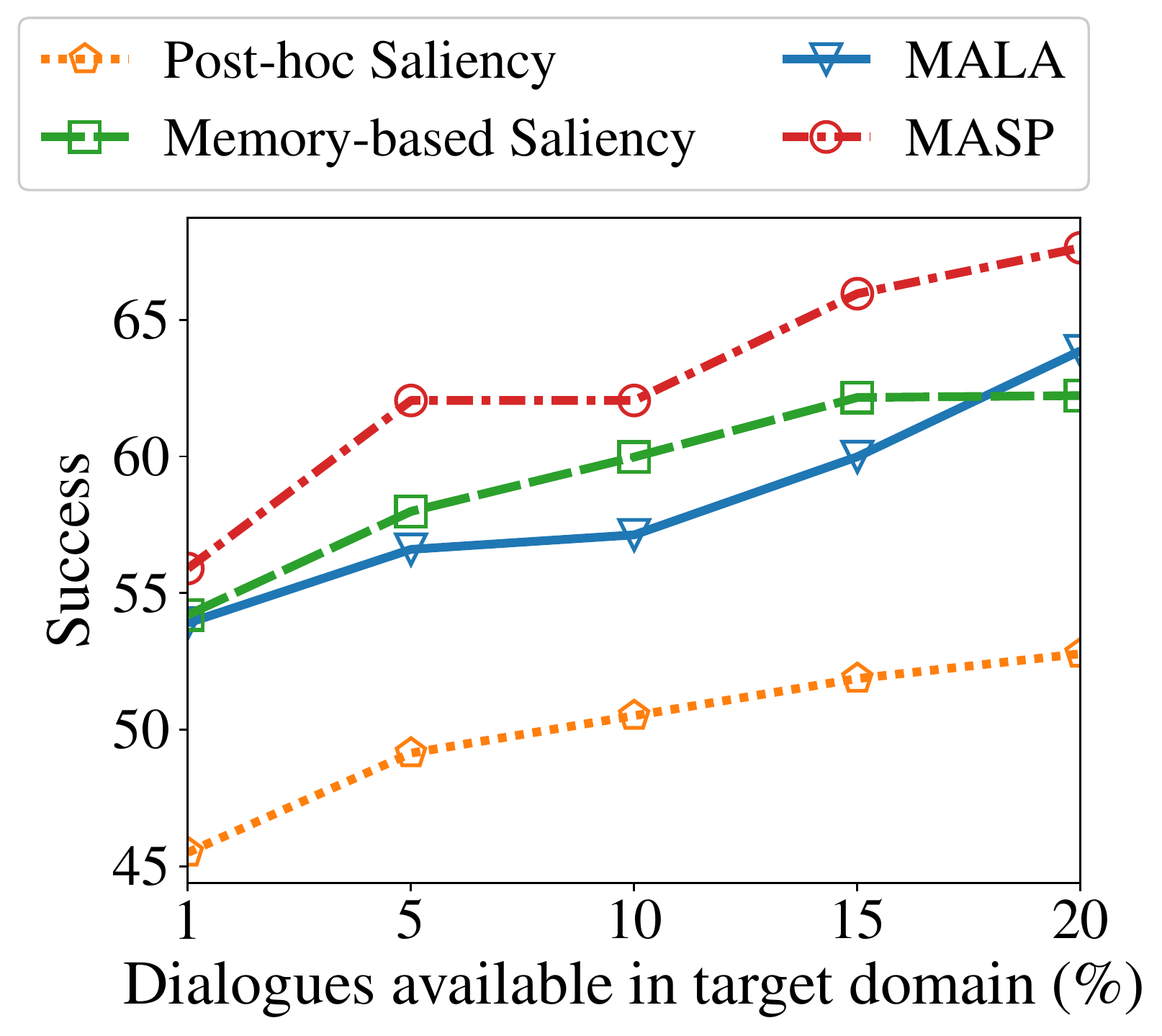}
\label{vrnn-FU}
\end{overpic}
}
\caption{Effects of action characteristics}
\label{abla-target-train}
\end{figure}

We first study the effects of different components of MASP in the cross-domain setting.
We compare MASP and its two variants with MALA (discrete action) under different dialogue ratios in target domains.
The results are shown in Fig. \ref{abla-target-attr} and Fig. \ref{abla-target-train}.
We can see that Memory-based Saliency is more comparable to MALA when using \textit{train} as target domain, especially when the dialogue ratio is low. 
This is largely because there are many shared knowledge of system intentions and state transitions between \textit{taxi} and \textit{train} domains, and the memory component could benefit from such knowledge via the dialogue state tracking model.
On the other hand, for target domains that do not have much advantage (e.g., \textit{attraction}), the pseudo parallel corpus might contribute more to action learning.  
This conclusion is also consistent with what we observe in multi-domain joint training.

Last, we study the effects of content planning model design.
We consider mainly two types of content planning model that works on natural language actions: action decoder and classifier, denoted as Act-$\textsc{Dec}$ and Act-$\textsc{Cls} $, respectively. 
Specifically, an action decoder generates a text span and feed it to the language generation model, while action classifier conducts classification to select one action from the action set given by the training set. 
We also consider to enhance the planning model with (1) action embeddings computed by summing word-embedding of words in actions; (2) memory component as discussed in \ref{conditional-resp}.
% of different content planning models are
From the results shown in Table \ref{tab-action-model}, we can see that reusing the memory component could effectively improve the  performance of conditioned response generation.
% For example, 
We also find that action classifier generally perform better than action decoder, while the latter is more flexible to manipulate the content to generate. 
This is aligned with our intuition since more specific and task-relevant intentions are more favorable for task-oriented dialogues.
% intentions that are more specific and task-relevant are more favorable for task-oriented dialogues.  

Moreover, through natural language actions, we could obtain a transparent response generation process, where the decided intermediate action is human-understandable.
% (see appendix for case studies)
% Natural language actions provide an essential benefit of enabling interpretable response generation, which is a key prerequisite of trustworthy dialogue systems.
% From the perspective of practicality, interpretability
Such transparency could help alleviate the credit assigning issue by identifying the effectiveness of dialogue planning and surface realization. 
% despite lack of fine-grained annotations.
Table \ref{app-case-study} shows that the proposed approach can obtain interpretable action representations (e.g., "request-departure") for the utterances that have the same intention but with different wording. This table also shows an error that our approach made in action learning, where the sentence highlighted in bold expresses "inform-address" instead of "inform-area". This might be caused by that the utterance contains multiple intentions and is thus more challenging for action learning. 
Table \ref{app-case-generate} shows that, with the learned natural language actions, we can better identify the source of errors in conditioned response generation. The two generated responses read naturally sound but express inappropriate intentions. The upper and lower examples showcase an action decision error and a language generation error, respectively. These help recognize the cause of errors and guide further optimization of the relevant components (content planning model or surface realization model).

\begin{table*} [tbp]
    \centering
\begin{threeparttable}
    \centering
\caption{ \normalsize Natural Language Action Learning Results.
} \label{app-case-study}
% \tiny
%\scriptsize
%\footnotesize
\small
% \normalsize  
\setlength\tabcolsep{3.0pt}
\begin{tabular}{L{3.55cm}|L{11.0cm}}
\toprule
% \multicolumn{2}{c}{\textbf{System utterances}} \\
\textbf{Natural language action} & \textbf{System utterances} \\
\midrule
\multirow{3}{*}{\makecell[l]{\{\textit{`request', `departure'}\}} } & there are many trains during that time , where are you leaving from ? \\
\cmidrule{2-2}
& i am sorry , to help narrow down the results please reply with where you will be departing from \\
\cmidrule{2-2}
& i am going to need a little more information from you . where will you be leaving from ? \\
\cmidrule{1-2}
\multirow{3}{*}{\makecell[l]{\{\textit{`inform', `price', `area',} \\\textit{`offer', `reservation'}\}} } & \textbf{the address is 169 high street chesterton} and the price range is fairly expensive . would you like to make reservations ? \\ 
\cmidrule{2-2}
& i have one that is called saigon city . it 's more expensive and located in the north . can i make a reservation for you ? \\ 
\cmidrule{2-2}
&    we have 14 indian restaurants in the expensive category . do you have any more information to narrow down the search ? \\
\bottomrule
\end{tabular}
\begin{tablenotes}\small
\item[*] The parts highlighted in bold is the missing information of action representation. 
\end{tablenotes}
\end{threeparttable}
\end{table*}

\begin{table*} [!ht]
    \centering
\begin{threeparttable}
    \centering
\caption{ \normalsize Better Credit Assigning via Natural Language Actions
} \label{app-case-generate}
% \tiny
%\scriptsize
%\footnotesize
\small
% \normalsize  
\setlength\tabcolsep{3.0pt}
\begin{tabular}{L{3.55cm}|L{11.0cm}}
\toprule
% \multicolumn{2}{c}{\textbf{System utterances}} \\
User Utterance & i would like to leave from kings lynn sometime after 10:00 . \\
\midrule 
Ground-truth Response & we have 14 trains leaving after 10:00 , what time would you prefer to arrive by ? \\
\cmidrule{1-2}
Selected Action & \{\textit{`inform', `leaveat', \textbf{`request', `departure}'}\}   \\
\cmidrule{1-2}
Generated Response & i have train leaving after 10:00, \textbf{where would you like to depart? }\\
\midrule
\midrule
User Utterance & is there a restaurant in the center serving italian then ? \\
\midrule 
Ground-truth Response & there are several italian restaurants downtown , would you like me to pick for you ? \\
\cmidrule{1-2}
Selected Action & \{\textit{`inform', `food', `area'} \} \\
\cmidrule{1-2}
Generated Response & there are several italian restaurants , \textbf{do you have a preference to the area ?} \\

\bottomrule
\end{tabular}
\begin{tablenotes}\small
\item[*] The parts highlighted in bold is the error of either content planning or language generation. 
\end{tablenotes}
\end{threeparttable}
\end{table*}

\section{Related Work}
% Conditioned response generation aims to generate meaningful and fluent responses via intermediate meaning representations (i.e., actions). 
Early studies of conditioned response generation focus on enriching the meaning representations in task-oriented dialogues, e.g., utilizing graph structures and hierarchies among actions \cite{chen2019semantically,shiquanyang2020}, decomposing into fine-grained actions \cite{shu-etal-2019-modeling}, or encoding syntax attributes \cite{balakrishnan-etal-2019-constrained}. 
Since these approaches often assume expensive action annotations, recent years have seen a growing interest in learning latent actions in an unsupervised way \cite{zhao2019rethinking,huang2019mala}.
These approaches build on either adversarial learning \cite{Hu2017TowardCG,wang2018kdgan, yang2018unsupervised} or variational inference \cite{Kingma2014AutoEncodingVB} and encode all system utterances via a self-reconstruction task or distant supervision \cite{yarats2018hierarchical}. 
Due to their implicit nature, latent actions are difficult to generalize, and we aim to overcome this limitation by learning explicit action representations.

Our study is also related to attribution approaches, which aims to find features or regions of input that are important for tasks. 
Different types of techniques, including gradient-based \cite{Selvaraju2017GradCAMVE} and post-hoc \cite{Ribeiro2018AnchorsHM}, are applied for reinforcement learning \cite{Mott2019TowardsIR}, computer vision \cite{Adebayo2018SanityCF}, and text classification \cite{Jin2020TowardsHI}.
While these works focus on interpreting model behaviors, we aim to find salient words beyond input and utilize them as action representations.

\section{Conclusions}
We propose explicit action learning to achieve generalizable and interpretable dialogue generation. 
Our proposed model MASP learns unified and compact action representations.    
We propose a memory component that summarizes system utterances into natural language actions, i.e., spans of words from a unified vocabulary. 
We further introduce an auxiliary task to encourage natural language actions to only preserve task-relevant information.
Experimental results confirm that MASP achieves better performance compared with the state-of-the-art in different settings, especially when supervision is limited. 
We plan to consider structural action representation learning that could convey more information as future work.

\section*{Acknowledgement}
We would like to thank Xiaojie Wang for the insightful discussions.
This work is supported by Australian Research Council (ARC) Discovery Project DP180102050.

% \newpage
\clearpage

\bibliographystyle{acl_natbib}
\bibliography{emnlp2020}

\end{document}